# Continuous Value Function Approximation for Sequential Bidding Policies

**Craig Boutilier**
Dept. of Computer Science
University of British Columbia
Vancouver, BC V6T 1Z4
*cebly@cs.ubc.ca*

**Moisés Goldszmidt and Bikash Sabata**
SRI International
333 Ravenswood Ave
Menlo Park, CA 94025
*{moises, sabata}@erg.sri.com*

## Abstract

Market-based mechanisms such as auctions are being studied as an appropriate means for resource allocation in distributed and multiagent decision problems. When agents value resources in combination rather than in isolation, they must often deliberate about appropriate bidding strategies for a sequence of auctions offering resources of interest. We briefly describe a discrete dynamic programming model for constructing appropriate bidding policies for resources exhibiting both complementarities and substitutability. We then introduce a continuous approximation of this model, assuming that money (or the numeraire good) is infinitely divisible. Though this has the potential to reduce the computational cost of computing policies, value functions in the transformed problem do not have a convenient closed form representation. We develop *grid-based* approximations for such value functions, representing value functions using piecewise linear approximations. We show that these methods can offer significant computational savings with relatively small cost in solution quality.

## 1 Introduction

A great deal of attention has been paid to the development of appropriate models and protocols for the interaction of agents in distributed and multiagent systems (MASs). Often agents need access to specific resources to pursue their objectives, but the needs of one agent may conflict with those of another. A number of market-based approaches have been proposed as a means to deal with resource allocation and related problems in MASs [6, 20, 22]. Of particular interest are *auction mechanisms*, where each agent bids for a resource according to some protocol, and the allocation and price for the resource are determined by specific rules [14]. Auctions have a number of desirable properties as a means for coordinating activities, including minimizing the communication between agents and, in some cases, guaranteeing Pareto efficient outcomes [14, 22].

In order to effectively make use of market mechanisms, an agent must be aware of the resources it needs, their value, and how best to obtain them. In sequential decision making under uncertainty, an agent will typically consider a number of potential courses of action and settle on one with highest expected utility. However, when different courses of action require different collections of resources to be implemented, an agent must develop rational bidding strategies in order to obtain the most desirable resource sets. In many cases, these resources will be made available by different sellers at different times and at uncertain prices. As a consequence, optimal bidding behavior in a *sequence of auctions* is of considerable interest. Of course, similar considerations apply to other forms of market interaction as well: what resources should be purchased, at what prices, at what time, what portion of the budget should be set aside for specific resources, and so on.

In the setting described above, an agent often requires several resources (a resource *bundle*) before pursuing a particular course of action. Obtaining one resource without another—for example, being allocated trucks without fuel or drivers, or processing time on a machine without skilled labor to operate it—makes that resource worthless. Such resources are said to exhibit *complementarities*. Furthermore, resource bundles are generally *substitutable*: obtaining the bundle needed to pursue one course of action can lower the value of obtaining another, or render it worthless. For instance, once trucks and drivers are obtained for transporting material in an optimal fashion, helicopters and pilots lose any value they may have had.

Complementarities and substitutability complicate the process of bidding on resources. A key difficulty that arises in the sequential model is how an agent computes bids for individual resources. An agent has a valuation for a particular resource bundle $b = \{r_1, \cdots, r_k\}$, but has no independent assignment of value to the individual resources. While auction theory can tell us how an agent should bid as a function of its valuation of resource $r_i$ for specific auction mechanisms [14], in our setting no such valuation exists. If $b$ is worth $v(b)$, how is an agent to "distribute the value" among the resources $r_i$ in order to compute bids?[1]

In earlier work [3], we described a sequential auction model and a dynamic programming algorithm for constructing optimal bidding policies when agents valued bundles that exhibited the type of complementarity and substitutability that arises in sequential decision models. Specifically, assuming resources are auctioned in a known order, we modeled the bidding problem as a *Markov decision process* (MDP), and described how an agent could construct an optimal bidding policy for the sequence of auctions based on its valuations of different resource bundles. Agents can choose how much (and whether) to bid for a resource depending on past successes, failures, prices, and so on. Unfortunately, the number of bids available at any point in time is generally very large. Given any state $d$ of the agent's endowment (say, measured in dollars), the agent can bid any amount less than $d$ for the good in question. Since endowments can usually be divided quite finely, this induces very large state and action spaces, causing computational difficulty for discrete dynamic programming. While the complexity of the algorithm grows only linearly in the size of the endowment, large endowments, or endowments that are finely divisible, often cause greater computational difficulty than the number of resources under consideration, inducing MDPs with very large state and action spaces.

In this paper, we investigate continuous approximations of this model, allowing an agent's endowment and set of possible bids to vary continuously. While this clearly *expands* the state and action spaces, we do this in the hope of using continuous function maximization methods to choose optimal bids. Once again, difficulties arise because value functions in this model generally do not have a convenient closed form representation. To deal with this, we introduce *grid-based approximation methods* for computing value functions. In particular, we sample the value function at specific points in state space (at various endowment levels) and use linear interpolation to determine the value function at other points in state space. This is similar in spirit to the use of grid-based methods for approximating value functions (w.r.t. belief space) in partially observable Markov decision processes [4, 11, 13, 12]. We show that the piecewise linear (PWL) value functions constructed in this fashion approximate the true value functions for the sequential bidding problem quite closely in many instances, while requiring considerably less computational effort.

The remainder of the paper is organized as follows. In Section 2 we describe the basic resource allocation problem under consideration and review the MDP model of this problem and the DP algorithm for bidding policy construction of [3]. In Section 3 we describe a continuous approximation of the discrete MDP, where endowment is treated as a continuous component of state space and the action space (possible bids) is similarly treated continuously. We describe a *fixed-grid method* for approximating the value functions in the continuous MDP that constructs piecewise linear approximations of the value functions, and whose error can be bounded *a posteriori*. In Section 4 we describe empirical results using the fixed grid approximation. We show that computing value functions at a small number of sample points and interpolating can offer significant computational savings in constructing value functions and the induced policies, yet often provides very good approximations to both the optimal value function and optimal policy. We also demonstrate that, as expected, increased grid density offers better solution quality at higher computational cost, allowing anytime tradeoffs to be addressed within our model. In Section 5, we briefly describe two variable grid methods for value function approximation that introduce grid points into the approximation in places estimated to provide highest reduction in approximation error. We conclude in Section 6 with a discussion of related work and suggestions for future research.

## 2 The Discrete DP Model

### 2.1 Resource Allocation Problems

We assume we have a finite collection of agents, all of whom require resources from a pool of $n$ resources $R = \{r_1, \cdots, r_n\}$. We denote by $R^t$ the subset $\{r_1, \cdots, r_t\}, t \leq n$, with $R^0 = \emptyset$ by convention. We describe the quantities relevant to a specific agent $a$, since our focus in the paper is on the computation of policies for a fixed agent. Agent $a$ can use exactly one bundle $b^i = \{r_1^i, \cdots, r_{|b^i|}^i\}$ of resources from a set of $k$ *possible bundles*: $B = \{b^1, \cdots, b^k\}$. We denote by $U = \cup B$ the set of *useful resources* for our agent. Generally, $a$ need only worry about the properties (e.g., expected prices and demand) of resources in this set.[2] For this reason, we take $U$ to be identical to $R$ (possibly by ignoring irrelevant resources in $R$).

Agent $a$ has a positive valuation $v(b^i)$ for each resource bundle $b^i \in B$. This may, for instance, reflect the expected value of some course of action which requires the resource in $b^i$. Suppose the *holdings* of $a$, $h \subseteq R$, are those resources it is able to obtain. The value of these holdings is given by $v(h) = \max\{v(b^i) : b^i \subseteq h\}$; that is, the agent will be able to use the resource bundle from among those it holds in entirety with maximal value, with the others going unused. This is consistent with our interpretation given in Section 1 where resource bundles correspond to alternative plans for achieving some objective.[3] We denote by $\mathcal{H}$ the set of possible holdings (i.e., $\mathcal{H} = 2^R$).

---

[1] Complementarities are often addressed under the guise of combinatorial and simultaneous auctions; we discuss these briefly in Section 6.

[2] This is not true when demand for elements of $U$ is correlated with that for other resources. We do not address this possibility in this paper (see [3]).

[3] Our model can be extended to deal with bundles that stand in more general subadditive relations than the complete substitutability described here (though all such relations could be represented in the form described here with simple transformations). We keep this assumption for ease of presentation.

The resources $R$ of interest to $a$ will be auctioned off sequentially in a fixed, known order: without loss of generality, we assume that this ordering is $r_1, r_2, \cdots, r_n$. We use $A_i$ to denote the auction for $r_i$. Agent $a$ is given an initial endowment $e$ of some common numeraire good (we'll use dollars) which it can use to obtain resources. At the end of the round, $a$ has holdings $h$ and $d$ dollars remaining from its endowment.[4] We assume that the utility of being in such a state at the end of the round is given by $v(h) + f(d)$, where $f$ is some function attaching utility to the unused portion of the endowment. Other utility functions could be considered, but this form is often suitable.

There are a wide range of options one could consider when instantiating this framework, with regard to the type of auctions used, the information provided to agents, and so on (see [3] for more on this). We assume that the individual auctions will be first-price, sealed-bid—each agent will provide a single bid and the highest bidder will be awarded the resource for the price bid. We adopt this model because of the ease with which it fits with our sequential model for bid computation; however, we believe our model could be adapted to other auction protocols as well as to other forms of market interaction. We also assume that bids are discrete (integer-valued): that is, bids are not arbitrarily divisible. Additionally, we assume that agents, once obtaining a resource, cannot resell that resource to another agent. This, of course, means that an agent may obtain one resource $r_i$, but later be unable to obtain a complementary resource $r_{i+k}$, essentially being "stuck" with a useless resource $r_i$. We do this primarily for simplicity, though in certain settings this assumption may be realistic. We are currently exploring more sophisticated models where agents can "put back" resources for re-auctioning, or possibly resell resources directly to other agents. Finally, we assume that agent $a$ believes that the highest bid that will be made for resource $r_i$, excluding any bid $a$ might make, is drawn from some unknown distribution $Pr^i$. Because bids are integer-valued, this unknown distribution is a multinomial over a non-negative, bounded range of integers.[5]

We make two remarks on this model. First, if the space of possible bids is continuous, a suitable continuous PDF (e.g., Gaussian) could be used to model bid distributions (and uncertainty about the parameters of this PDF, if necessary). Second, we make an implicit assumption that bids for different resources are uncorrelated. By having independent distributions $Pr^i$ rather than a joint distribution over *all* bids, agent $a$ is reasoning as if the bids for different resources are independent. When resources exhibit complementarities, this is unlikely to be the case. For instance, if someone bids up the price of some resource $r_i$ (e.g., trucks), they may subsequently bid up the price of complementary resource $r_j$ (e.g., fuel or drivers). If agent $a$ does not admit a model that can capture such correlations, it may make poor bids for certain resources. Once again, we make this assumption primarily for ease of exposition. Admitting correlations does not fundamentally change the nature of the algorithms to follow, though it does raise interesting modeling and computational issues [3].

### 2.2 Computing Bids by Dynamic Programming

The difficulty in computing bids for the sequential auctions $A_i$ lies in the fact that the agent does not have a specific valuation for any individual good; rather, it has valuations over bundles. This suggests an agent should compute a *bidding policy* in which bids for specific resources are conditioned on the outcomes of previous auctions. In [3] we model this problem as a fully observable MDP [16, 2]. The computation of an optimal bidding policy can be implemented using a standard stochastic dynamic programming algorithm such as value iteration. We briefly recap this model below.

As we will point out later in this section, optimal policy construction may be computationally intensive. The main goal of this paper to examine specific approximations to ease this burden. However, this dynamic programming model deals with the complementarities and substitutability inherent in our resource model; no special devices are required. Furthermore, it automatically deals with issues such as uncertainty, dynamic valuation, "sunk costs," and so on. Finally, we stress that given stationary, uncorrelated bid distributions, the computed policy is optimal.

We assume the decision problem is broken into $n+1$ stages, $n$ stages at which bidding decisions must be made, and a terminal stage at the end of the round. We use a time index $0 \leq t \leq n$ to refer to stages—time $t$ refers to the point at which auction $A_{t+1}$ for $r_{t+1}$ is about to begin. The *state* of the decision problem for agent $a$ at time $t$ is given by two variables: $h^t \subseteq R^t$, the subset of resources $R^t$ held by agent $a$; and $d^t$, the dollar amount (unspent endowment) available for future bidding. We write $\langle h, d \rangle^t$ to denote the state of $a$'s decision problem at time $t$.

The dynamics of the decision process can be characterized by $a$'s estimated transition distributions. Assuming that prices are drawn independently from the stationary distributions $Pr^i$, agent $a$ can predict the effect of any action (bid) $z$ available to it. If agent $a$ is in state $\langle h, d \rangle^t$ at stage $t$, it can bid for $r_{t+1}$ any amount $0 \leq z \leq d^t$. Letting $w$ denote the highest bid offered by other agents, if $a$ bids $z$ at time $t$, it will transition to state $\langle h \cup \{r_{t+1}\}, d - z \rangle^{t+1}$ with probability $Pr^{t+1}(w < z)$ and to $\langle h, d \rangle^{t+1}$ with $Pr^{t+1}(w \geq z)$.[6]

The final piece of the MDP is a reward function $q$. We simply associate a reward of zero with all states at stages 0 through $n - 1$, and assign reward $v(h) + f(d)$ to every terminal state $\langle h, d \rangle^n$. A *bidding policy* $\pi$ is a map-ping from states into actions: for each legal state $\langle h, d\rangle^t$, $\pi(\langle h, d\rangle^t) = z$ means that $a$ will bid $z$ for resource $r_{t+1}$. The *value* $V^\pi(\langle h, d\rangle^t)$ of policy $\pi$ at any state $\langle h, d\rangle^t$ is the expected reward $E_\pi(q(\langle h, d\rangle^n)|\langle h, d\rangle^t)$ obtained by executing $\pi$. The expected value of $\pi$ given the agent's initial state $\langle \emptyset, e\rangle^t$ is simply $V^\pi(\langle \emptyset, e\rangle^t)$. An *optimal bidding policy* is any $\pi$ that has maximal expected reward at every state.

---

[4] If speculation or reselling is allowed, there is the possibility that $d > e$, depending on the interaction protocols we allow. We ignore this possibility here.

[5] We assume that some reasonable bound can be placed on the highest bid. In [3] we represent $a$'s uncertainty over the parameters of this distribution with a Dirichlet distribution. The expected values of these parameters can be used here without difficulty.

[6] For expository purposes, the model assumes ties are won. Several rules can be used for ties; none complicate the analysis.

We compute the optimal policy using value iteration [16], defining the value of states at stage $t$ using the value of states at stage $t + 1$. Specifically, we set

$$V(\langle h, d\rangle^n) = v(h) + f(d)$$

and define, for each $t < n$:

$$\begin{aligned}
Q(\langle h, d\rangle^t, z) &= \Pr^{t+1}(w < z) \cdot V(\langle h \cup \{r_t\}, d - z\rangle^{t+1}) \\
&\quad + \Pr^{t+1}(w \geq z) \cdot V(\langle h, d\rangle^{t+1}) \quad (1)\\
V(\langle h, d\rangle^t) &= \max_{z \leq d} Q(\langle h, d\rangle^t, z) \quad (2)\\
\pi(\langle h, d\rangle^t) &= \arg\max_{z \leq d} Q(\langle h, d\rangle^t, z) \quad (3)
\end{aligned}$$

With $V$ defined for all stage $t + 1$ states, $Q(\langle h, d\rangle^t, z)$ denotes the value of bidding $z$ at state $\langle h, d\rangle^t$ and acting optimally thereafter. $V(\langle h, d\rangle^t)$ denotes the optimal value at state $\langle h, d\rangle^t$, while $\pi(\langle h, d\rangle^t)$ is the optimal bid.

Implementing value iteration requires that we enumerate, for each $t$, all possible stage $t$ states and compute the consequences of every feasible action at that state. This can require substantial computational effort. While linear in the state and action spaces (and in the number of stages $n$), the state and action spaces themselves are potentially quite large. The number of possible states at stage $t$ could potentially consist of any subset of resources $R^t$ together with any monetary component. The action set at a state with monetary component $d$ has size $d + 1$. Fortunately, we can manage some of the complexity associated with various resource combinations by using certain pruning and generalization strategies; a number of these are described in [3].

Reducing the impact of the number of possible bids is more difficult. We can certainly restrict the state and action space to dollar values no greater than $a$'s initial endowment $e$. If the PDF is well-behaved (e.g., concave), pruning is possible: for instance, once the expected value of a larger bids starts to decrease, search for a maximizing bid can be halted. Another method for dealing with this is to assume that endowment and bids are continuous. We turn our attention to this strategy in the next section.

We point out that the model described above does not allow for strategic reasoning on the part of the bidding agent. The agent takes the expected prices as given and does not attempt to compute any form of equilibrium. The motivation for this model is described in more detail in [3]: briefly, we assume there that the price models will be adjusted over time with the aim of converging to some form of equilibrium. We expect that the MDP model described here could be extended to allow for equilibrium computation. However, there are several reasons for using the approach described above rather than a full Bayes-Nash equilibrium model [21]. First, equilibrium computation is often infeasible, especially in a nontrivial sequential, multi-resource setting like ours. Second, the information required on the part of each agent, namely a distribution over the possible *types* of other agents, is incredibly complex—an agent type in this setting is its valuations for *all* resource bundles, making the space of types unmanageable in general.[7] Finally, the common knowledge assumptions usually required for equilibrium analysis are unlikely to hold in this setting. Our model is thus more akin to limited rationality models (e.g., fictitious play [5, 9]), in particular, when agents attempt to learn bid distributions over time [3]. A consequence of this approach is that, in early rounds, allocations may not be efficient. However, learning behavior tends to lead to efficient outcomes after some number of rounds, and often leads to optimal allocations (with respect to social welfare) [3].

## 3 Continuous Approximations of Bids and Endowments

When an agent's initial endowment is large or finely-divisible, value iteration can be computationally expensive. For instance, given an endowment of $10,000 which can be bid in $1 increments, the state space of the MDP has size $10,000 \cdot |\mathcal{H}|$ and the action space has size 10,000. An endowment of $100 with penny increments is just as large. In order to deal with the computational complexity incurred by such endowments, we assume that endowments and bids are continuous-valued. This allows us to use continuous optimization methods to compute optimal bids as a function of the state and represent our value functions in a continuous fashion. In Section 3.1 we describe the continuous MDP, while in Section 3.2 we present a fixed-grid, approximate representation for value functions, and show how to solve the MDP using this representation.

### 3.1 Continuous Version of the MDP

Viewing money as continuous requires that we make the following adjustments to the MDP described in Section 2. At any state where the agent's remaining endowment is $d$, the agent can consider bids in the interval $[0, d]$; and given a maximum endowment $m$, state space ranges over any endowment value in the interval $[0, m]$. Note that the state space $S = \mathcal{H} \times [0, m]$ has both a discrete and continuous component. Since bids are now continuous, we assume the agent models the high bid distributions using a continuous density function. We generally assume that simple parametric distributions or mixtures are used for this purpose (e.g., in Section 4, we use Gaussian bid distributions).

Value functions in the discrete MDP presented in the last section are represented as a table of values. With the hybrid MDP—containing both continuous and discrete components—we represent a value function $V^t$ as a table of continuous functions. For each resource holding $h \subseteq R^t$, we define the one-dimensional, continuous function $V_h^t(d) = V^t(\langle h, d \rangle)$; that is, $V_h^t$ describes how $V^t$ varies with remaining endowment for a fixed set of holdings $h$.

---

[7] We use *type* here in the sense used in game theory for games with incomplete information [15].

In order to implement DP with the continuous dimension in the state space, we require some manageable representation for the continuous value function components $V_h^t$. At stage $n$ (i.e., with zero stages to go), $V_h^n(d)$ is simply equal to $q(\langle h, d \rangle)$; since the reward function is specified in a suitable form, $V_h^n$ will have a tractable form for each $h \subseteq R$. For instance, if our agent has linear utility for remaining endowment—that is, if $q(\langle h, d \rangle) = v(h) + \alpha d$—then $V_h^n$ can be represented with two parameters. Constructing $V_h^{n-1}$ requires that we backup values through possible bids (actions) as in Equation (1) and then choose the bid with highest Q-value to determine $V_h^{n-1}$ as in Equation (2).

For any specific state $\langle h, d \rangle$, the computation of $V_h^{n-1}(d)$ is not problematic. Equation (2) requires that we find the bid $z$ that maximizes $Q(\langle h, d \rangle^{n-1}, z)$. As long as the reward function is monotonically increasing with remaining endowment and the probability of winning a resource increases with higher bids (and both are well-behaved), it is easy to see that the Q-function has a unique maximum. Thus the corresponding constrained maximization problem is easy to solve. The difficulty lies in the fact that we cannot compute the value function at the infinitely many states (varying with possibly endowment). In general, the value function $V_h^{n-1}$ will not have a convenient closed form, nor will $V_h^t$ for any $t < n$.

To circumvent this difficulty, we adopt a grid-based approach to approximating the continuous (component) value functions at each stage. This method of approximation is fairly common in representing value functions in continuous-state MDPs. Continuous domains are studied quite frequently in reinforcement learning (RL), and arise in the conversion of POMDPs to belief state MDPs. In fact, the grid-based approach to computing value functions for POMDPs is a commonly used approximation technique [4, 11, 13, 12]. Unlike belief state MDPs (which have an $n - 1$-dimensional state space, where $n$ is the number of system states) or many multi-dimensional control and RL problems, our domain has only one continuous dimension. As such, we are not affected by the curse of dimensionality that often plagues grid-based approximation methods in other areas: increasing the density of our grid causes only a linear increase in required computational effort.

### 3.2 A Uniform Fixed Grid Approach

Let us assume that we are given some representation of the continuous value function components $V_h^t$ for each $h \subseteq R^t$. Our grid-based approaches to computing an approximation of $V_h^{t-1}$ all work as follows. First, a set of *grid points* is chosen: these form a small sample of the possible remain-ing endowments $d$ at stage $t - 1$. These grid points are chosen over the interval $[0, m]$ (where $m$ is the maximum initial endowment), and we assume that both 0 and $m$ are among the set. Let there be $g$ such points: $0 = d_1 < d_2 < \cdots < d_g = m$. At each such point $d_i$, we compute $V_h^{t-1}(d_i)$ using Equation (2). We note again that this is a fairly routine continuous maximization problem, and that $Q^{t-1}$ is constructed using the given functions $V_h^t$. If each $V_h^t$ is correct, then we have computed the exact value function $V_h^{t-1}$ at these grid points. We define $V_h^{t-1}$ over the entire endowment range using linear interpolation. For any endowment $d$, let $d_j \leq d \leq d_{j+1}$. We set

$$\widetilde{V}_h^{t-1}(d) = V_h^{t-1}(d_j) + \frac{V_h^{t-1}(d_{j+1}) - V_h^{t-1}(d_j)}{d_{j+1} - d_j}(d - d_j)$$

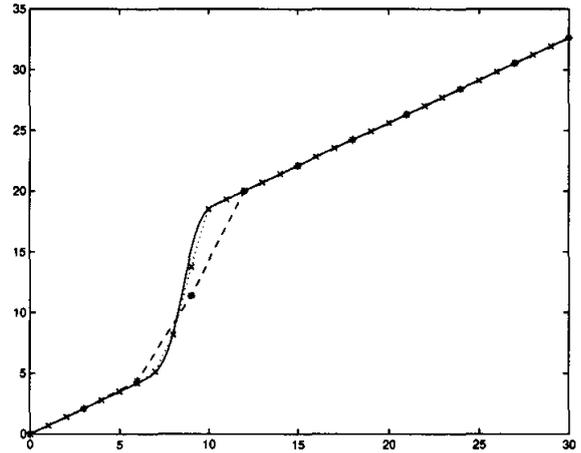

Figure 1: Uniform grid approximation: the solid line represents the true value function, the dashed line an approximation using a coarse grid, and the dotted line an approximation using a fine grid.

The interpolation process is illustrated in Figure 1 for two different grid granularities. Note that $\widetilde{V}_h^{t-1}(d)$ is an approximation to the true value $V_h^{t-1}(d)$ in the hybrid MDP.

Our uniform, fixed grid approach to value function computation assumes that a fixed grid is specified in advance, with $g$ grid points uniformly covering the interval $[0, m]$. This grid is not adjusted during computation, nor does it vary across stages or across different states (with different holdings $h$). Specifically, the steps above are repeated at each stage: $V_h^t(d_i)$ is computed at each grid point $d_i$ with respect to the approximate value functions $\widetilde{V}_h^{t+1}$; and $\widetilde{V}_h^t$ is defined at all remaining points by interpolation over the values $V_h^t(d_i)$ at the grid points.

The grid-based approach allows one to determine *a posteriori* bounds in the error in the value function. Suppose that we are given an accurate $t+1$-stage value function $V^{t+1}$ (so that $V_h^{t+1}(d)$ is correct for all $h$ and all $d$). Because we com-

86  Boutilier, Goldszmidt, and Sabata

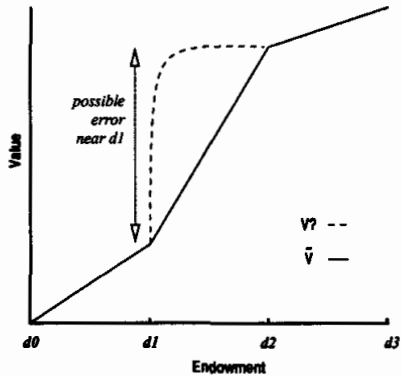

Figure 2: Maximum error introduced by uniform grid approximation: the solid line represents the linear approximation; the dashed line is one possible, ill-behaved, true value function consistent with it.

pute $V_h^t(d_i)$ exactly at each grid point $d_i$, we are assured that the resulting approximate value function $\widetilde{V}_h^t$ is correct at each of these grid points (for all $h$). Now consider the difference in value $\delta_i^h = V_h^t(d_{i+1}) - V_h^t(d_i)$ between any two consecutive grid points $d_i$ and $d_{i+1}$. Since the value function is monotonically increasing with endowment (for any reasonable utility function), this difference is positive. Furthermore, the error in the approximate value function $\widetilde{V}_h^t$ on the interval $[d_i, d_{i+1}]$ must be bounded by this difference $\delta_i^h$. To see this, see the extreme value function candidate illustrated in Figure 2 between grid points $d_1$ and $d_2$. For any $h \subseteq R^t$, define $\delta^h = \max_{i<g} \delta_i^h$, and for any stage $t$ define $\delta(V^t) = \max_{h \subseteq R^t} \delta^h$. This maximum value difference bounds the error in the estimated value function $\widetilde{V}^t$:

**Proposition 1** *Let approximate value function $\widetilde{V}^t$ be generated (for all $h \subseteq R^t$) with a fixed grid approach, using the exact value function $V^{t+1}$. Then $\|\widetilde{V}^t - V^t\| < \delta(V^t)$.*

Of course, we generally construct $\widetilde{V}^t$ using an *approximation* $\widetilde{V}^{t+1}$ of $V^{t+1}$, thus the errors can accumulate; but they do so in an additive fashion.

**Proposition 2** *If the continuous MDP for sequential auctions (with n stages) is solved using a fixed grid approximation, then*

$$\|\widetilde{V}^t - V^t\| < \sum_{i=t+1}^{n} \delta(V^i)$$

*for all $t \leq n$.*

We note that these bounds, while tight theoretically, are generally not reached in practice, as we shall see in the next section. Error as large as this can only be reached for value functions that behave very badly. Finally, using standard arguments regarding how error in value functions manifests itself in behavioral error, we note that the difference in value of the (greedy) policy induced by the approximate value function and the optimal policy is bounded by twice the error in the value function.

## 4 Empirical Evaluation

We have implemented the dynamic programming algorithm for sequential auctions in Matlab, and have experimented with the approximation of value functions using sampled, continuous functions. We performed 20 experiments each consisting of four runs. Each experiment comprises the computation of a bidding policy for an agent requiring four bundles of resources drawn from a set of ten potentially useful resources. The number of resources per bundle was generated from a Gaussian distribution with mean 3 and variance 1. The valuation for the bundles was also generated from a Gaussian with mean 15 and variance 2, while the estimated bid distributions were Gaussians with means in the range of 3 to 6 for different resources and variance of 0.5. The utility function used is $v(h) + 0.7d$ (so remaining endowment is valued at 70 cents to the dollar).[8]

The four runs in each experiment differ on the number of samples used in the approximation. One of the runs, which we identify as Discrete, consists of an initial endowment of $30, with bids that can be incremented discretely in $1 units. The dynamic program computing the bidding policy of the agent is based on the algorithm described in section 2.2. Since there are no approximations involved, this run is our "gold standard" against which we compare the approximations produced by the fixed grid methods. The optimal value function at the initial state in the Discrete model (i.e., expected value obtained in the sequence of auctions) tended to lie between 25 and 33 for the 20 different trials.[9] The remaining three runs consist of the fixed grid strategy with continuous bids, using 5, 10, and 15 sample points, and are denoted as G5, G10, and G15, respectively.

We use several pruning techniques to reduce the number of states (i.e., different resource combinations considered at different stages). This influences both the computation time and the reporting of the errors. Since in a large number of these states the best policy is to bid zero (e.g., given the current resources it will be impossible to complete a bundle with the resource being auctioned), considering all of these states in error computation conveys misleading average error statistics (our approximations would look too good). By eliminating these states, we report on a more meaningful measure of error between the optimal result and our approximations. In addition, we report relative error results rather than absolute error (scaling relative to the magnitude of the true value at each point). Errors reported are squared differences using $\left(\frac{e-v}{v}\right)^2$, where $e$ is the estimated value using

---

[8] Discounting endowment is simply a convenient way to raise the relative values of all bundles uniformly.

[9] This is higher than the expected value of not competing for resources, which has a value of 21 (= 0.7 · 30).

|           | Number of States | Mean Squared Value Error | Mean Maximum Squared Value Error | Mean Squared Policy Error | Mean Maximum Squared Policy Error |
|-----------|------------------|--------------------------|----------------------------------|---------------------------|-----------------------------------|
| Discrete  | 14415            | 0                        | 0                                | 0                         | 0                                 |
| 5 Grid    | 2790             | 0.2698                   | 3.9008                           | 0.2204                    | 3.0842                            |
| 10 Grid   | 5115             | 0.1508                   | 2.4330                           | 0.0650                    | 0.9408                            |
| 15 Grid   | 7440             | 0.0169                   | 0.2966                           | 0.0367                    | 0.5202                            |

Table 1: Aggregated results over all the states and all experiments.

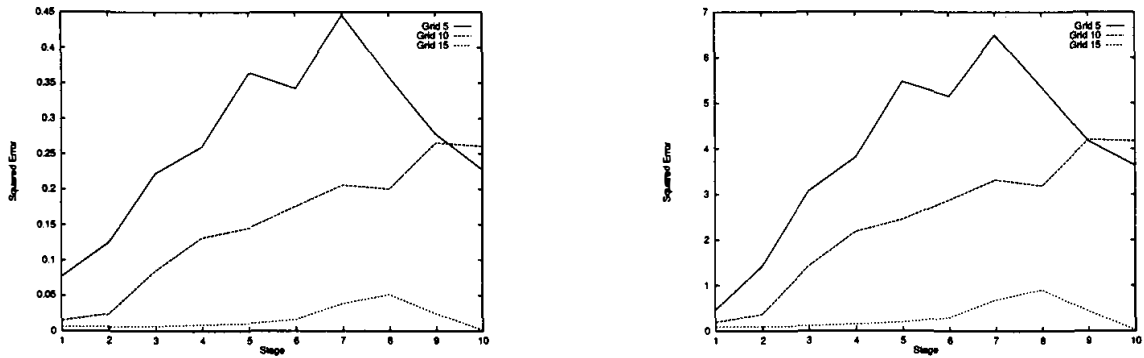

Figure 3: Mean (left) and Maximum (right) Value Error Per Stage

the approximation and $v$ is the true value. If $v < 1$, we do not normalize.

Table 1 provides the summary of the number of states explored by each method, the mean error in the value function, the mean error in the "optimal" policy induced by the corresponding value function, and the maximum error in the value function and induced policy. These are computed (averaged or maximized) over all (unpruned) states in all stages of the dynamic program. The numbers reported are the averages over the 20 experiments.

As can be seen, we have a reduction in error (both in the value function and the induced policy) as we increase the number of sample points in the grid. Also there is an expected (linear) increase in computational effort as we increase the number of sample points. These results strongly suggest that with a fraction of the computational effort (5 grid points) we can obtain decent approximations to the value function and the optimal policy. Notice also that the error in induced behavior is generally smaller than the error in the approximate value function. This is because the incorrect value function still induces optimal bids at many states. The number of states explored by the approximations is a fraction of the the number required by the original, accurate model, but the computation per stage is sometimes more. To display the average and maximum errors in the value function at different stages of the bidding process, we compute the average error over all (unpruned) states at each stage and over all 20 experiments. Figure 3 shows this stage-by-stage error (mean and maximum) in the value function, while Figure 4 shows the stage-by-stage error in the induced approximate policies.

## 5 Variable Grid Approaches

One difficulty with the fixed grid approach is that computational effort is sometimes spent computing values at grid points that do not improve the accuracy of the PWL approximation of the value function, while candidate grid points that could reduce the error substantially are ignored. In the experiments described above, this was sometimes observed to be the case. For this reason we consider several *variable grid* strategies which introduce grid points dynamically based on some measure of the likely improvement they will make in the value function estimate.

**Method VG1:** Based on the error analysis above, one way to ensure that the maximum error is reduced as much as possible with as few grid points as possible is to introduce grid points in those intervals that have the largest maximum error. Variable grid method VG1 does just this. We assume that we have a set of grid points $d_1, \cdots d_n$ whose values $V_h^t(d_i)$ have been computed. (Initially, we assume that we have grid points at endowment level 0 and endowment level $m$, the maximum). For each consecutive pair of grid points, the difference in value $V_h^t(d_{i+1}) - V_h^t(d_i)$ computed. A grid point is introduced at the midpoint between that pair of grid points whose value difference is the largest. This continues until some maximum number of grid points has been introduced or the maximum value difference between any pair of adjacent grid points falls below some threshold. We know that the error in our approximation $\widetilde{V}_h^t$ is bounded by

88     Boutilier, Goldszmidt, and Sabata

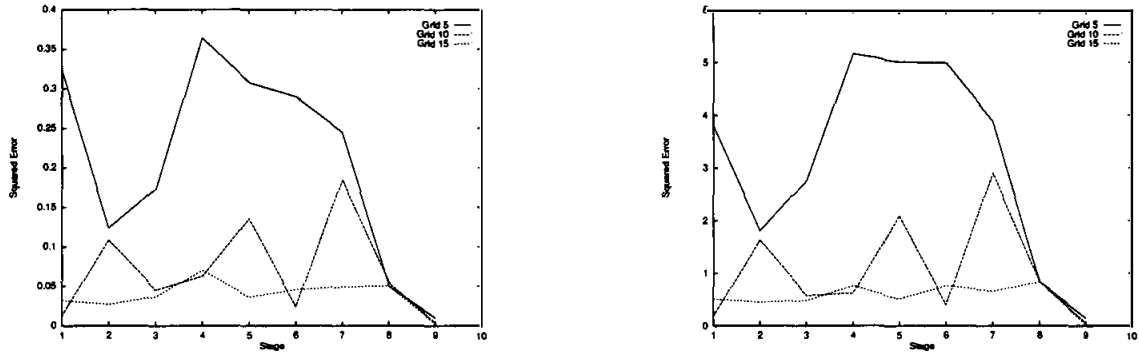

Figure 4: Mean (left) and Maximum (right) Policy Error Per Stage

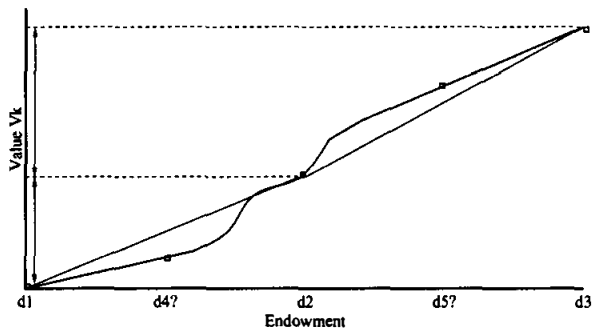

Figure 5: Variable Grid Method VG1

this maximum difference (as described in the previous section). This adaptive grid method improves this error bound the fastest fashion possible.

Figure 5 illustrates this method of introducing grid points. Here the grid point $d_5$ is introduced rather than $d_4$ since $V_h^t(d_3) - V_h^t(d_2)$ is larger than $V_h^t(d_2) - V_h^t(d_1)$. Choosing the maximum value difference can be implemented very quickly using a priority queue: the pair of adjacent grid points with the maximum value difference is popped off the queue, and the two new intervals created by the insertion of a new grid point are added to the queue using the value difference as a key.

**Method VG2:** One difficulty with method VG1 is that a lot of effort can be expended introducing grid points that have no real effect on the value function estimate. Specifically, if there is a segment of the value function that is actually linear, the introduction of a grid point in that region does not improve our estimate.[10] For instance, adding a new point between $d_5$ and $d_3$ in the example above might improve the guaranteed error bound by the largest amount. But since it lies on a linear segment, it doesn't reduce the *actual error* in the value function estimate at all. Introducing a point elsewhere (e.g., at $d_4$, or between $d_2$ and $d_5$) would be a more effective use of computational resources.

---

[10] Because we often adopt a linear utility component for remaining endowment at the end of the round of auctions (see Section 4), such linear segments arise rather frequently.

To capture this intuition, we say that the *error reduction* offered by the introduction of grid point $d_5$ is the absolute difference between $V_h^t(d_5)$ (which we computed when that grid point was introduced) and the previous predicted value for $d_5$ before the introduction of the point (that is, it's predicted value using the linear segment from $d_2$ to $d_3$). If the introduction of $d_5$ caused a small reduction in error, this suggests that introducing points on either side of it may not be useful in reducing error. Another method of measuring this is by examining the angles between the linear interpolates. If the angle is close to $180°$, then adding new points will not improve the accuracy significantly. The closer the angle is to $90°$, the greater the odds of improving the error by introducing new grid points.

Our second variable grid method, **VG2**, requires the introduction of *pairs* of grid points at each iteration. Suppose we have grid points $d_1, \cdots d_n$ whose values $V_h^t(d_i)$ have been computed. Whenever a grid point is introduced, we compute its *error reduction factor* (ERF): the difference between its computed value and the value it was predicted to have just before it was introduced as a grid point. We keep *unexpanded grid points* ordered in a priority queue, sorted according to their ERFs. When the grid point $d_i$ with the largest ERF is removed from the queue, it is considered *expanded*. Expanding a grid point $d_i$ requires that we introduce two new grid points: one that bisects the interval $[d_{i-1}, d_i]$; and one that bisects the interval $[d_i, d_{i+1}]$. These two new grid points are added to our value function and inserted into the priority queue for subsequent expansion.

This second method works extremely well when value functions have large linear segments. In Figure 6, we see that after the introduction of $d_4$ and $d_5$ surrounding $d_3$, the error reduction factor of $d_4$ is close to zero, since it lies on a linear segment between $d_3$ and $d_2$. The ERF for $d_5$ is shown by the arrows, and is greater; so we introduce points $d_6$ and $d_7$ surrounding $d_5$, ignoring further splitting of the intervals around $d_4$. Notice that we can "fool" this method. If we in-troduce a grid point $d_8$ between $d_7$ and $d_2$ (in the middle of the "S-shaped" portion of the value function), we compute a very small ERF for $d_8$; this is misleading since new grid points surrounding $d_8$ would be very useful. While potentially problematic, such circumstances seem to arise rarely, at least in preliminary experimental evaluation.

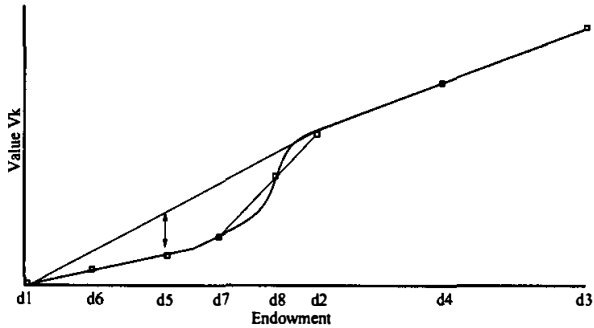

Figure 6: Variable Grid Method VG2

One could introduce a number of other variable grid generation techniques. While the two methods described above are intuitively appealing and easy to implement (and have low computational cost), others may be well-suited to different types of problems or different utility functions. We have not yet experimented with these techniques in depth. We note that the error bounds derived for the fixed grid method above apply directly to variable grid methods as well. Given a set of variable grid points, the error in the value function approximation can be bounded by the $\delta_i$ factors (the maximum differences in values at consecutive grid points).

# 6 Concluding Remarks

## 6.1 Related Work

Auctions involving complementary goods have been studied widely, though it is unknown whether simple selling mechanisms can lead to efficient outcomes [1, 22]. Two methods for dealing with complementarities have been studied in some depth in the literature: simultaneous auctions for multiple goods [1, 18]; and *combinatorial auctions* in which agents submit bids for *resource bundles* [17, 19, 22]. Neither of these models is suitable in the setting we consider, when resources are made available at different points in time or are offered by different sellers.

Even in settings where the requirements of combinatorial or simultaneous auctions are met—or could be enforced—a sequential model has some attractive features. Unlike combinatorial models, it relieves the (computational) burden of determining a final allocation from the seller, effectively distributing computation among the buyers (as in the simultaneous case). This can be important, as determining an optimal allocation (maximizing the seller's revenue) is NP-hard [19]. Our sequential model also has the advantage that buyers are not required to reveal information about their valuations for specific resource bundles that they do not obtain. Furthermore, it has greater flexibility in that agents can enter and leave the market without forcing recomputation of entire allocations. In contrast to simultaneous models, agents in the sequential model lessen their exposure. If an agent does not obtain a certain resource early in the sequence, it need not expose itself by bidding on complementary resources occurring later in the sequence. Agents are typically bidding in a state of greater knowledge in the sequential model, at least during later auctions.

While bidding strategies for sequential auctions would seem to be an issue worthy of study, there appears to have been little research focused on this issue. What work exists (see, e.g., [8, 10]) tends to focus the seller's point of view—for example, will simultaneous or sequential sales maximize revenue—and does not address the types of complementarities and substitutability we consider here.

## 6.2 Summary and Future Directions

In this paper we reviewed a model for computing optimal sequential bidding policies for resources that exhibit complementarities and substitutability, and described a continuous approximation of the model. We presented several grid-based approximate representations for value functions, and described a dynamic programming algorithm that uses these PWL representations. While the resulting policies may not be optimal, we provided error bounds on the value functions and policies produced, and showed empirically that the fixed-grid method works quite well: it produces high quality value functions and policies with a small portion of the computational effort required by an exact algorithm. We also illustrated the "contract anytime" flavor of these algorithms: with denser grids (thus, more computational effort) one can produce higher quality policies.

We have not yet experimented with the variable grid approaches in detail. We expect these algorithms to outperform the fixed grid algorithm, especially in domains where linear utility for remaining endowment is adopted—this is due to the frequent linear segments exhibited by certain parts of the value function. It is also the case the general strategy of adopting PWL approximations is especially appropriate when these linear utility fragments abound. We hope to explore other approximate representations that are suitable for other "typical" utility functions.

There are a number of other issues we hope to explore in the near future. One is the appropriate modeling of correlations in prices. As mentioned above, when goods exhibit complementarities, it is highly unlikely that prices will be uncorrelated. Modeling this simply requires that the agent maintain a joint distribution reflecting price expectations.[11] The difficulty lies in strategy computation—when one price is observed, expected prices for future resources may change, requiring a change in "planned" behavior. In other words, the decision problem is truly partially observable and requires some form of history-dependent policy [3].

---

[11] Clearly, representations that exploit a certain amount of independence can be used as well.

Another avenue we hope to explore is the integration of the adaptive model explored in [3]—where prices are learned during multiple rounds of auctions—with the continuous approximation strategy used here. Note that to include adaptivity in the continuous model described in Section 3 requires only the update of the bid distributions, which for simple parametric forms and mixture models (e.g., Gaussians) is a well-known process [7].

Finally, we want to explore the deeper issues involved with integrating an agent's "object-level" sequential reasoning (i.e., the decision making in which the courses of action that consume resources are developed) with the type of market reasoning described here. Specifically, we envision the emergence of very interesting policy patterns; for example, an agent might decide not to bid for resources until it executes part of its plan because the uncertainty associated with that plan's outcome may make obtaining the resource too risky until it is more certain of the plan's outcome. A "compound MDP" modeling all levels of reasoning seems to be an appropriate framework for thinking about such problems.

## Acknowledgements

Craig Boutilier and Moisés Goldszmidt acknowledge partial support from the DARPA Co-ABS program (through Stanford University contract F30602-98-C-0214). Bikash Sabata was funded by DARPA through the SPAWARSYSCEN under Contract Number N66001-97-C-8525. Craig Boutilier was partially supported by NSERC Research Grant OGP0121843. We would like to thank Bill Walsh for his helpful comments, suggestions and pointers to relevant literature, as well as Yoav Shoham, Tuomas Sandholm and Mike Wellman for their comments and pointers.

## References

[1] S. Bikhchandani and J. W. Mamer. Competitive equilibria in and exchange economy with indivisibilities. *J. of Economic Theory*, 74:385–413, 1997.

[2] C. Boutilier, T. Dean, and S. Hanks. Decision theoretic planning: Structural assumptions and computational leverage. *J. Artif. Intel. Research*, 1999. To appear.

[3] C. Boutilier, M. Goldszmidt, and B. Sabata. Sequential auctions for allocation of resources with complementarities. *IJCAI-99*, Stockholm, 1999. To appear.

[4] R. I. Brafman. A heuristic variable-grid solution method for POMDPs. *AAAI-97*, pp.727–733, Providence, 1997.

[5] G. W. Brown. Iterative solution of games by fictitious play. In T. C. Koopmans, ed., *Activity Analysis of Production and Allocation*. Wiley, New York, 1951.

[6] S. Clearwater, ed. *Market-based Control: A Paradigm for Distributed Resource Allocation*. World Scientific, San Mateo, 1995.

[7] R. O. Duda and P. E. Hart. *Pattern Classification and Scene Analysis*. Wiley, New York, 1973.

[8] R. Engelbrecht-Wiggans and R. J. Weber. A sequential auction involving assymetrically informed bidders. *Intl. J. Game Theory*, 12:123–127, 1983.

[9] D. Fudenberg and D. K. Levine. *The Theory of Learning in Games*. MIT Press, Cambridge, MA, 1998.

[10] D. B. Hausch. Multi-object auctions: Sequential vs. simultaneous sales. *Management Science*, 32(12):1599–1610, 1986.

[11] M. Hauskrecht. A heuristic variable-grid solution method for POMDPs. *AAAI-97*, pp.734–739, Providence, 1997.

[12] C. C. White III and W. T. Scherer. Solutions procedures for partially observed Markov decision processes. *Operations Research*, 37(5):791–797, 1989.

[13] W. S. Lovejoy. Computationally feasible bounds for partially observed Markov decision processes. *Operations Research*, 39(1):162–175, 1991.

[14] R. P. McAfee and J. McMillan. Auctions and bidding. *J. Econ. Lit.*, 25:699–738, 1987.

[15] R. B. Myerson. *Game Theory: Analysis of Conflict*. Harvard University Press, Cambridge, 1991.

[16] M. L. Puterman. *Markov Decision Processes: Discrete Stochastic Dynamic Programming*. Wiley, 1994.

[17] S. J. Rassenti, V. L. Smith, and R. L. Bulfin. A combinatorial auction mechanism for airport time slot allocation. *Bell J. Econ.*, 13:402–417, 1982.

[18] M. H. Rothkopf. Bidding in simultaneous auctions with a constraint on exposure. *Op. Res.*, 25:620–629, 1977.

[19] M. H. Rothkopf, A. Pekeč, and R. M. Harstad. Computationally manageable combinatorial auctions. *Mgmt. Sci.*, 1998. To appear.

[20] T. Sandholm. Limitations of the vickrey auction in computational multiagent systems. *ICMAS-96*, pp.299–306, Kyoto, 1996.

[21] W. Vickrey. Counterspeculation, auctions, and competitive sealed tenders. *J. Finance*, 16(1):8–37, 1961.

[22] M. P. Wellman, W. E. Walsh, P. R. Wurman, and J. K. MacKie-Mason. Auction protocols for decentralized scheduling. (manuscript), 1998.